%% file: main.tex
\documentclass[11pt,a4paper]{article}
\usepackage[hyperref]{acl2021}
\usepackage{times}
\usepackage{latexsym}

\usepackage[colorinlistoftodos]{todonotes}
\usepackage{microtype}
\usepackage{microtype}
\usepackage{rotating}
\usepackage{blindtext}
\usepackage{amsmath}
\usepackage{amsfonts}
\usepackage{graphicx}
\usepackage{mathtools}
\usepackage{booktabs}
\usepackage{relsize}
\usepackage{float}
\usepackage{graphics}
\usepackage{multirow}
\usepackage{algorithm}
\usepackage{algpseudocode}
\usepackage[utf8]{inputenc}

\DeclareUnicodeCharacter{01B0}{\`u}
\DeclareUnicodeCharacter{01A1}{\`o}

\usepackage{threeparttable}
\usepackage{booktabs, caption, makecell}

\usepackage[title]{appendix}
\usepackage[cal=cm]{mathalfa}
\usepackage{multirow}
\usepackage[export]{adjustbox}
\usepackage{subcaption}
\usepackage{pifont}

\aclfinalcopy 
\def\aclpaperid{***} 

\newcommand{\mycomment}[1]{}                     
\newcommand{\ignore}[1]{}

\usepackage{xspace}

\usepackage{amsmath}
\usepackage{amsfonts}
\usepackage{xcolor}

\title{The DCU-EPFL Enhanced Dependency Parser at the 
IWPT 2021 Shared Task}

\author{James Barry\textsuperscript{1}, \hspace{0.1cm} Alireza Mohammadshahi\textsuperscript{2, 3}, \hspace{0.1cm} Joachim Wagner\textsuperscript{1}, \hspace{0.1cm} Jennifer Foster\textsuperscript{1}, \\ {\bf James Henderson\textsuperscript{2}}  \\
$^{1}$ ADAPT Centre, School of Computing, Dublin City University \\
$^{2}$ Idiap Research Institute \\
$^{3}$ École Polytechnique Fédérale de Lausanne---EPFL \\
$^{1}$ \texttt {\{james.barry, joachim.wagner, jennifer.foster\}@adaptcentre.ie} \\
$^{2}$ \texttt {\{alireza.mohammadshahi, james.henderson\}@idiap.ch}
}

\date{}

\begin{document}
\maketitle

\begin{abstract}
\input{abstract.tex}
\end{abstract}

\section{Introduction}
\label{sec:intro}
\input{intro.tex}

\section{Related Work}
\label{sec:related}
\input{related.tex}

\section{Official System Overview}
\label{sec:overview}

\input{overview.tex}

\section{Experiments}
\label{sec:experiments}
\input{experiments.tex}
\section{Conclusion}
\label{sec:conclusion}
\input{conclusion.tex}

\ignore{***

\section{Supplementary Materials}
\label{sec:supplementary}

Supplementary material may report preprocessing decisions, model parameters, and other details necessary for the replication of the experiments reported in the paper.
Seemingly small preprocessing decisions can sometimes make a large difference in performance, so it is crucial to record such decisions to precisely characterize state-of-the-art methods. 

\subsection{Appendices}
\label{sec:appendix}
Appendices are material that can be read, and include lemmas, formulas, proofs, and tables that are not critical to the reading and understanding of the paper. 
In the submitted version, appendices should be uploaded as a separate file.
\paragraph{\LaTeX-specific details:}
In your camera ready version use {\small\verb|\appendix|} before any appendix section to switch the section numbering over to letters.

\subsection{Software and Data}
\label{sec:software_and_data}
Submissions may include software and data used in the work and described in the paper.
Any accompanying software and/or data should include licenses and documentation of research review as appropriate.

***}

\section*{Acknowledgments}
This research is supported by Science Foundation
Ireland through the ADAPT Centre for Digital Content Technology, which is funded under the SFI Research Centres Programme (Grant 13/RC/2106)
and is co-funded under the European Regional Development Fund. This research is also supported by the Swiss National Science Foundation, grant CRSII5-180320.
We would like to thank the shared task organizers as well as the anonymous reviewers for their helpful feedback.

\bibliographystyle{acl_natbib}
\bibliography{acl2021, anthology}


\end{document}

%% file: abstract.tex
We describe the \textit{DCU-EPFL} submission to the
\emph{IWPT 2021 Shared Task on Parsing into Enhanced Universal Dependencies}.
The task involves parsing Enhanced UD graphs,
which are an extension of the basic dependency trees
designed to be more facilitative towards representing semantic structure.
Evaluation is carried out on 29 treebanks in 17 languages
and participants are required to parse the data from each language starting from raw strings.
Our approach uses the Stanza pipeline to preprocess the text files,
XLM-RoBERTa to obtain contextualized token representations,
and
an edge-scoring and labeling model to predict the enhanced graph.
Finally, we run a post-processing script to ensure all of our outputs are valid Enhanced UD graphs.
Our system places 6th out of 9
participants
with a coarse Enhanced Labeled Attachment Score (ELAS) of 83.57.
We carry out additional post-deadline experiments which include using Trankit for pre-processing,
XLM-RoBERTa\textsubscript{LARGE}, treebank concatenation,
and multitask learning between a basic and an enhanced dependency parser.
All of these modifications improve our initial score and our final system has a coarse ELAS of 88.04.

%% file: intro.tex
The \emph{IWPT 2021 Shared Task on Parsing into Enhanced Universal Dependencies} \citep{bouma-etal-2021-overview}
is the second task involving the prediction of Enhanced Universal Dependencies (EUD) graphs\footnote{\url{https://universaldependencies.org/u/overview/enhanced-syntax.html}} following the 2020 task 
\citep{bouma-etal-2020-overview}.
EUD graphs are an extension of basic UD trees, designed to be more useful in shallow natural language understanding tasks \citep{schuster-manning-2016-enhanced} and lend themselves more easily to the representation
of semantic structure than strict surface structure dependency trees.
In the shared task,
the enhanced graphs must be predicted from raw text, i.e.\ 
participants
must segment the input into sentences and tokens.
Participants are encouraged to predict
lemmas, Part-of-Speech (POS) tags, morphological features and basic dependency trees
as well.

Our system, \textit{DCU-EPFL}, uses a single multilingual Transformer \citep{NIPS2017_3f5ee243} encoder,
namely
XLM-RoBERTa (XLM-R) \citep{conneau-etal-2020-unsupervised},
which is a multilingual RoBERTa model \citep{liu2019roberta}, to obtain contextualized token encodings.
These are then passed to 
the enhanced dependency parsing model.
The system is straightforward to apply to new languages with
enhanced UD annotations.
In the official submission, we use the same hyper-parameters
for all languages.
Our parsing component can produce arbitrary graphs,
including graph structures where words may have multiple heads
and cyclic graphs.
Our system uses the following three components:

\begin{enumerate}
    \item Stanza \citep{qi-etal-2020-stanza} for sentence segmentation, tokenization and the prediction of all UD features apart from the enhanced graph.
    \item A Transformer-based dependency parsing model to predict Enhanced UD graphs.
    \item A post-processor ensuring that every graph is a rooted graph where all nodes are reachable from the notional root token.
\end{enumerate}

Our official system placed 6th out of 9
teams
with a coarse Enhanced Labeled Attachment Score (ELAS) of 83.57.
In a number of unofficial post-evaluation experiments,
we make
four
incremental changes to our pipeline approach:
\begin{enumerate}
    \item We replace the Stanza pre-processing pipeline with Trankit
          \citep{nguyen-etal-2021-trankit}.
    \item We use XLM-R\textsubscript{Large} instead of
          XLM-R\textsubscript{Base}.
    \item We concatenate treebanks from the same language which have more
          than one training treebank and concatenating English treebanks
          to the Tamil training data.
    \item We introduce a novel multitask model which parses the basic UD
          tree and enhanced graph in tandem.
\end{enumerate}
All of these additional steps improved our evaluation scores,
and for our final system, which incorporates all additional modifications,
our evaluation score increases from 83.57 to 88.04.
Our code is publicly available.\footnote{\url{https://github.com/jbrry/IWPT-2021-shared-task}}

%% file: related.tex
In this section, we discuss the relevant literature related to Enhanced Universal Dependencies.

\subsection{Enhanced Universal Dependencies}

Despite
the recent wave of Deep Learning models and accompanying analyses that show that such models learn information about syntax, there is still interest and merit in utilizing hierarchically structured representations such as trees and semantic representations to provide greater supervision about what is taking place in a
sentence \cite{oepen-etal-2019-mrp}.
While dependency trees are often used in downstream applications,
their structural restrictions may hinder the representation of content words \citep{schuster-manning-2016-enhanced}.
The Enhanced UD representation tries to fill this gap by enabling more expressive graphs in the UD format, which capture phenomena such as
added subject relations in control and raising,
shared heads and dependents in coordination,
the insertion of null nodes for elided predicates,
co-reference in relative clause constructions
and augmenting modifier relations with
prepositional or case-marking information.

\newcite{schuster-manning-2016-enhanced} build on the Stanford Dependencies (SD) initiative \citep{Marneffe:2006}
and extend certain flavors of the SD dependency graph representations to UD in the form of enhanced UD relations for English.
They use a rule-based system that converts basic UD trees to enhanced UD graphs based on 
dependency structures identified to require enhancement.
\newcite{nivre-etal-2018-enhancing} use rule-based and data-driven approaches in a cross-lingual setting
for bootstrapping enhanced UD representations in Swedish and Italian and show that both techniques are capable of annotating enhanced dependencies in different languages.

\subsection{The IWPT 2020 Shared Task on Parsing Enhanced Universal Dependencies}

The first shared task on parsing Enhanced Universal Dependencies \citep{bouma-etal-2020-overview} brought renewed attention to the problem of predicting enhanced UD graphs.
Ten teams submitted to the task.
The winning system \citep{kanerva-etal-2020-turku}
utilized the UDify model \citep{kondratyuk-straka-2019-75},
which uses a BERT model \citep{devlin-etal-2019-bert} as the encoder with multitask classifiers for POS-tagging, morphological prediction and dependency parsing built on top.
They developed a system for encoding the enhanced representation into the basic dependencies so it can be predicted in the same way as a basic dependency tree but with enriched dependency types that can then be converted into the enhanced structure.
In an unofficial submission shortly after the task deadline,
\newcite{wang-etal-2020-enhanced} outperform the winning system
using
second-order inference methods with Mean-Field
Variational Inference.

Most systems used pretrained Transformers to obtain token representations,
either by using the Transformer directly
\citep{kanerva-etal-2020-turku, grunewald-friedrich-2020-robertnlp, he-choi-2020-adaptation}
or passing the encoded representation to BiLSTM layers
where they are combined with other features such as
context-free
FastText word embeddings \citep{wang-etal-2020-enhanced},
character features
and features obtained from predicted POS tags, morphological features and
basic UD trees \citep{barry-etal-2020-adapt},
or are used as frozen embeddings
\citep{hershcovich-etal-2020-kopsala}.
The only transition-based system among the participating teams
\cite{hershcovich-etal-2020-kopsala}
used a stack-LSTM architecture \citep{dyer-etal-2015-transition}.
\newcite{ek-bernardy-2020-much} and \newcite{dehouck-etal-2020-efficient} combine basic dependency parsers and a rule-based system to generate EUD graphs from the predicted trees.

%% file: overview.tex
\begin{figure}
  \centering
  \includegraphics[width=\linewidth]{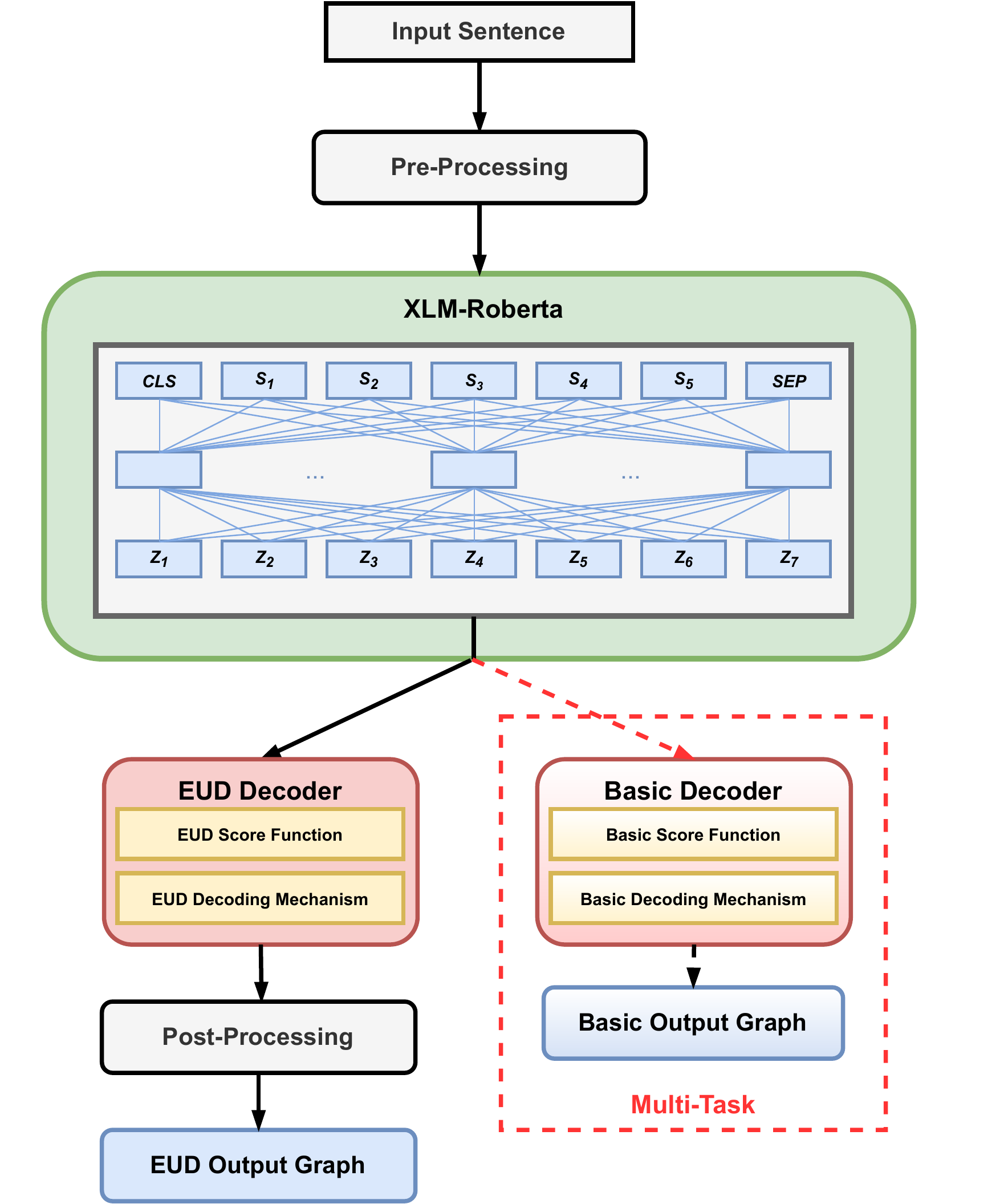}
  \caption{DCU-EPFL Architecture.}
  \label{fig:architecture}
\end{figure}

This section describes our official system, which is the system we submitted prior to the competition deadline.
The architecture of our system is shown in Figure~\ref{fig:architecture}.\footnote{For the official system, we did not include the basic dependency parser in a multitask setup.}
The raw text test files for each language contain a mixture of test data covering multiple treebanks, so participants do not know their exact domain.
For our official system, we choose the model trained on the treebank with the most amount of training data
in terms of sentences
for each language to process the test files.
This heuristic corresponds to using
Czech-PDT for Czech,
Dutch-Alpino for Dutch,
English-EWT for English,
Estonian-EDT for Estonian
and Polish-PDB for Polish.

\subsection{Pre-processing}

For sentence segmentation, tokenization
and the prediction of the base UD features
(all UD features apart from the enhanced dependency graphs and miscellaneous items in CoNLL-U files),
we use the Stanza library \citep{qi-etal-2020-stanza} trained on version 2.7 of the UD treebanks
for each treebank released as part of the training data for the shared task.\footnote{For Arabic,
    our Stanza Multi-word Token (MWT) expander predicted MWTs with a span of length 1 for two sentences.
    In the UD guidelines, MWT span lengths must be larger than one.
    To pass validation, we trained a UDPipe tokenizer \citep{straka-strakova-2017-tokenizing}
    with Word2Vec embeddings for Arabic
    instead.
}
Note that our parser
does not pre-suppose any input features other than the input text
but we predict the base features using our pre-processing pipeline for completeness
and to enable possible additional post-processing which involves altering enhanced dependency labels with lemma information.

\subsection{Enhanced UD Parsing}

For the enhanced UD parser, we use a Transformer encoder in the form of XLM-R \citep{conneau-etal-2020-unsupervised}
with a first-order arc-factored model which utilizes the edge and label scoring method of \citep{kiperwasser-goldberg-2016-simple}.
In initial experiments, we found this
model to perform better than biaffine attention \citep{DBLP:journals/corr/DozatM16} for the task of EUD parsing.
This finding
was also made by \citep{lindemann-etal-2019-compositional} and \citep{straka-strakova-2019-ufal} for the task of semantic parsing
across numerous Graphbanks \citep{oepen-etal-2019-mrp}.
\newcite{straka-strakova-2019-ufal} suggest that biaffine attention may be less suitable
for predicting whether an edge exists between any pair of nodes using a predefined threshold and is perhaps more suited for dependency parsing, where 
words are competing with one another to be classified as the head in a softmax layer.
The consistency of these findings across EUD and semantic parsing Graphbanks
may provide evidence
that enhanced UD is closer to semantic dependency parsing than basic UD parsing.

\paragraph{Parser Implementation}

Given a sentence $x$ of length $n$, our model computes
vector representations $\mathbf{R} = (\mathbf{r_1},\mathbf{r_2},...,\mathbf{r_n}$)
for the predicted tokens ($x_1,x_2,...,x_n$).
Since the WordPiece tokenization~\cite{wu2016googles} of XLM-R differs from the tokenization used in UD,
we track the mapping $I$ from XLM-R's $k$-th sub-word unit of the $j$-th input token
produced by Stanza to the sub-word unit's position $I_{j,k}$ in context of the sentence
and
we consider the output vector $\mathbf{e_{I_{j,1}}}$ of the first sub-word unit
of each word $x_j$ as its vector representation~($\mathbf{r}_j$):
\begin{align}
\begin{split}
    &\mathbf{E}=\mathrm{XLMR}(x_1, x_2, \dots, x_n) \\
    &\mathbf{R} = \mathrm{Filter}(E,\mathbf{I}) 
\end{split}
\end{align}
where
$\mathbf{E}=(\mathbf{e}_1,...,\mathbf{e}_N)$
are the output vectors of all sub-word units,
$N$ being the total number of subword units in the sentence,
and $\mathrm{Filter}()$ chooses the first embedding for each token.
We add a dummy representation of the same dimensionality
for the ROOT token to the sequence of vectors $\mathbf{R}$ but mask out predictions from this token.
Following \newcite{kiperwasser-goldberg-2016-simple},
these representations~$\mathbf{R} = (\mathbf{r_1},\mathbf{r_2},...,\mathbf{r_n}$)
are then passed 
to the dependency parsing component,
where the feature function $\phi$ is the concatenation of 
the representations of
a potential head word $x_h$ and dependent word $x_d$, where $ \circ $ denotes concatenation:
\begin{align}
    \phi(h, d)&=\mathbf{r_h} \circ \mathbf{r_d}
\end{align}

\paragraph{Edge Prediction}

We compute scores for all $n(n-1)$ potential edges $(h,d)$, $h \neq d$, with an MLP:
\begin{align}
    s_{hd}\textsuperscript{(arc)} = MLP\textsuperscript{(arc)}(\phi(h, d)) 
\end{align}
The edge classifier computes scores for all possible head-dependent pairs,
and we compute a sigmoid on the resulting matrix of scores to obtain probabilities.
We use an edge prediction threshold of 0.5, i.e.\ we include all edges
with a score above 0.5 in the preliminary EUD graph.
This enables words to have multiple heads but it can also lead
to words receiving no head,
where we manually select the edge that has the highest probability,
and to fragmented graphs, see
post-processing in Section~\ref{post-proc}.

\paragraph{Label Prediction}

To label the graph, we then choose a label for each edge
using a separate classifier:
\begin{align}
    s_{hd}\textsuperscript{(label)} = MLP\textsuperscript{(label)}(\phi(h, d)) 
\end{align}
The scores for all possible labels are passed to a softmax layer, which outputs the probability of each label for edge $(h,d)$ and we select the label with the highest probability for each edge.

\paragraph{Loss Function}
For edge prediction, sigmoid cross-entropy loss is used,
and for label prediction,
as we want to select the label for each chosen edge,
softmax cross-entropy loss is used \citep{dozat-manning-2018-simpler}.
We interpolate between the loss given by the edge classifier and the loss given by the label classifier \citep{dozat-manning-2018-simpler, wang-etal-2020-enhanced}
with a constant $\lambda$:

\begin{equation}
    \mathcal{L}=\lambda\mathcal{L}^{(label)}+(1-\lambda)\mathcal{L}^{(edge)}
\end{equation}

\paragraph{Training details}
For the empty nodes which are prevalent in enhanced UD graphs,
we added them into the graph,
and offset the head indices
to account for the new token(s) added to the graph.
At test time, we did not predict whether an elided token should be added to the graph.
Due to time constraints,
we trained using the full lexicalized enhanced dependency labels but intend to devise a delexicalization and relexicalization procedure in future work.

\begin{table}
\centering
\begin{tabular}{lr}
\hline \textbf{Hyperparameter} & \textbf{Size} \\ \hline
XLM-R\textsubscript{Base} Hidden Size & 768 \\
XLM-R\textsubscript{Large} Hidden Size & 1024 \\
\hline
Edge Feedforward & 300 \\
Label Feedforward & 300 \\
Input Dropout & 0.35  \\
Dropout & 0.35  \\
Edge Prediction Threshold & 0.5 \\
Loss interpolation $\lambda$ & 0.10  \\
\hline
\end{tabular}
\caption{\label{hyperparams} Hyperparameters of our EUD parsing model. }
\end{table}

\subsection{Post-processing}
\label{post-proc}

In the Enhanced UD guidelines,
the predicted structure must be a connected graph
where all nodes are reachable from the notional root\footnote{In UD, the notional ROOT is the token with ID 0, whereas a root node is any node that has 0 as its head.}.
After predicting the test files,
we use the graph connection tool in \citep{barry-etal-2020-adapt}
to make sure that each sentence is a connected graph.
Specifically, we repeatedly check for unreachable nodes and the number of unreachable nodes that can be reached from them.
We choose the candidate which maximises this number (in the case there are ties, we choose the first node in surface order) and makes it a child of the notional ROOT,
i.e.\ this node becomes an additional root node.
System outputs are then validated at level 2 by the UD validator \footnote{\url{https://github.com/UniversalDependencies/tools/blob/master/validate.py}}
to catch bugs prior to submission.

%% file: experiments.tex
In this section, we discuss our official results
and then describe post-deadline experiments that improved our submission's score.
Model hyperparameters are listed in Table~\ref{hyperparams}.
The choice of XLM-R encoder (Base or Large) determines the hyperparameters of the
encoder part of our model.
In our official submission, we use XLM-R\textsubscript{Base}.
A dropout value of 0.35 is used for the input embeddings as well as for the encoder and MLP networks.
A loss interpolation constant $ \lambda $ of 0.1 is used as in \citep{wang-etal-2020-enhanced}.

\begin{table*}[t]
\centering
  \begin{adjustbox}{width=\linewidth}
  \begin{tabular}{lcccccccccc}
    \toprule
    \multirow{2}{*}{Language} &
      \multicolumn{1}{c}{[1]} && \multicolumn{1}{c}{[2]} && \multicolumn{1}{c}{[3]} &&
      \multicolumn{1}{c}{[4]} && \multicolumn{1}{c}{[5]} \\
      \cline{2-2} \cline{4-4} \cline{6-6} \cline{8-8} \cline{10-10}
     & \textbf{Official} && \textbf{[1]+Trankit} && \textbf{[2]+XLM-R\textsubscript{Large}} && \textbf{[3]+Concat} && \textbf{[4]+MTL}\\
\midrule
Arabic & 71.01 && 78.05(+24.2\%) && 79.51(+6.6\%) && - && 81.72(+10.8\%)\\
Bulgarian & 92.44 && 92.47(+0.4\%) && 93.26(+10.5\%) && - && 93.59(+4.9\%)\\
Czech & 89.93 && 90.28(+3.5\%) && 91.06(+8.1\%) && 91.43(+4.1\%) && 91.30(-1.5\%) \\
Dutch & 81.89 && 86.51(+25.5\%) && 87.67(+8.6\%) && 88.60(+7.5\%) && 89.51(+7.9\%) \\
English & 85.70 && 85.97(+1.8\%) && 86.94(+6.9\%) && 87.46(+3.9\%) && 87.28(-1.4\%) \\
Estonian & 84.35 && 84.54(+1.2\%) && 85.92(+8.9\%) && 86.68(+5.4\%) && 86.76(+0.6\%) \\
Finnish & 89.02 && 89.34(+2.9\%) && 90.79(+13.6\%) && - && 91.16(+4.1\%)  \\
French & 86.68 && 86.80(+0.9\%) && 89.12(+17.6\%) && - && 90.38(+11.6\%)  \\
Italian & 92.41 && 92.44(+0.4\%) && 93.35(+12.1\%) && - && 93.47(+1.8\%)  \\
Latvian & 86.96 && 86.85(-0.8\%) && 88.81(+14.9\%) && - && 89.18(+3.3\%)  \\
Lithuanian & 78.04 && 78.44(+1.8\%) && 82.09(+16.9\%) && - && 83.47(+7.7\%)  \\
Polish & 89.17 && 89.30(+1.2\%) && 90.20(+8.4\%) && 91.15(+9.7\%) && 90.46(-7.8\%) \\ 
Russian & 92.83 && 93.06(+3.2\%) && 93.95(+12.8\%) && - && 94.09(+2.3\%) \\
Slovak & 89.59 && 90.81(+11.7\%) && 92.33(+16.5\%) && - && 92.73(+5.2\%) \\
Swedish & 85.20 && 85.98(+5.3\%) && 88.10(+15.1\%) && - && 88.64(+4.5\%)  \\
Tamil & 39.32 && 40.64(+2.2\%) && 48.85(+13.8\%) && 61.14(+24.0\%) && 62.06(+2.4\%)\\
Ukrainian & 86.09 && 86.30(+1.5\%) && 89.44(+22.91\%) && - && 90.91(+13.9\%)\\
\hline
\textbf{Average} & 83.57 && 84.58(+6.2\%) && 86.55(+12.7\%) && 87.48(+6.9\%) && 88.04(+4.5\%) \\
\hline
\end{tabular}
  \end{adjustbox}
\caption{\label{citation-guide}
Evaluation scores on the official test data on the language-specific test files.
All runs after \textbf{Official} subsume \textbf{Trankit} pre-processing
and all runs after \textbf{Trankit} subsume the \textbf{XLM-R\textsubscript{Large}} model. All numbers inside the parentheses are calculated as the relative error reduction of each column and its corresponding previous column.
}
\label{tab:eval}
\end{table*}

\subsection{Official Submission}
For the official submission,
we use the Stanza pre-processing pipeline and our dependency parsing model with XLM-R\textsubscript{Base}.
The results are listed in 
column [1]
of Table~\ref{tab:eval}.
Our official submission placed 6th of 9 participants.
The overall scores submitted by each team are listed in Table~\ref{tab:team_eval}.
The scores of two teams are close to our overall score: Combo and Unipi placed 4th and 5th with scores of 83.79 and 83.64 compared to our score of 83.57.
This grouping is outperformed by the top three submissions
TGIF, ShanghaiTech and RobertNLP 
by a margin from 3.2 ELAS points (RobertNLP vs.\ Combo) to 5.7 ELAS points
(TGIF vs.\ DCU-EPFL).



\subsection{Trankit Pre-processing}

In a post-deadline experiment,
we replace the Stanza pre-processing pipeline (which uses Word2Vec and FastText embeddings as external input features and a BiLSTM encoder) with Trankit \citep{nguyen-etal-2021-trankit}, which uses the Transformer XLM-R as the encoder.
The results from adopting Trankit for sentence segmentation and tokenization are listed in column [2] of Table~\ref{tab:eval}.
We notice slight improvements for all languages,
with notable exceptions being Arabic, Dutch and Slovak,
where the better pre-processing accounts for a
24.2\%, 25.5\% and 11.7\% relative error reduction. 

\subsection{XLM-R\textsubscript{Large}}

Our next modification is to leverage the XLM-R\textsubscript{Large} model.
This model has roughly twice as many parameters as the XLM-R\textsubscript{Base} model used in our official submission.
The results 
for combining Trankit pre-processing and using XML-R\textsubscript{Large}
are listed in column [3] of Table~\ref{tab:eval}.
The larger capacity of the model translates to
large relative error reductions
particularly for Finnish, French, Latvian, Lithuanian, Swedish, Tamil and Ukrainian.
Given the improvements seen by adopting both Trankit for pre-processing and the larger XLM-R\textsubscript{Large} model,
we now incorporate these modifications into all further experiments.

\begin{table*}[t]
    \centering
    \begin{adjustbox}{width=\linewidth}
    \begin{tabular}{lccccccccc|cc}
    \toprule
    Language & \bf combo & \bf dcu-epfl & \bf fastparse & \bf grew & \bf nuig & \bf robertnlp & \bf shanghaitech & \bf tgif & \bf unipi & \bf off. reference & \bf our best run \\
    \midrule
    Arabic & 76.39 & 71.01 & 53.74 & 71.13 & 0.0 & 81.58 & \textbf{82.26} & 81.23 & 77.17 & 67.35 & \textcolor{blue}{81.72} \\
    Bulgarian & 86.67 & 92.44 & 78.73 & 88.83 & 78.45 & 93.16 & 92.52 & \textbf{93.63} & 90.84 & 85.81 & \textcolor{blue}{93.59} \\
    Czech & 89.08 & 89.93 & 72.85 & 87.66 & 0.0 & 90.21 & \textcolor{blue}{91.78} & \textbf{92.24} & 88.73 & 78.44 & 91.30 \\
    Dutch & 87.07 & 81.89 & 68.89 & 84.09 & 0.0 & 88.37 & 88.64 & \textbf{91.78} & 84.14 & 82.48 & \textcolor{blue}{89.51} \\
    English & 84.09 & 85.70 & 73.00 & 85.49 & 65.40 & 87.88 & 87.27 & \textbf{88.19} & 87.11 & 83.68 & \textcolor{blue}{87.28} \\
    Estonian & 84.02 & 84.35 & 60.05 & 78.19 & 54.03 & 86.55 & 86.66 & \textbf{88.38} & 81.27 & 76.86 & \textcolor{blue}{86.76} \\
    Finnish & 87.28 & 89.02 & 57.71 & 85.20 & 0.0 & 91.01 & 90.81 & \textbf{91.75} & 89.62 & 78.26 & \textcolor{blue}{91.16} \\
    French & 87.32 & 86.68 & 73.18 & 83.33 & 0.0 & 88.51 & 88.40 & \textbf{91.63} & 87.43 & 98.80 & \textcolor{blue}{90.38} \\
    Italian & 90.40 & 92.41 & 78.32 & 90.98 & 0.0 & 93.28 & 92.88 & \textcolor{blue}{93.31} & 91.81 & 80.20 & \textbf{93.47} \\
    Latvian & 84.57 & 86.96 & 66.43 & 77.45 & 56.67 & 88.82 & 89.17 & \textbf{90.23} & 83.01 & 79.32 & \textcolor{blue}{89.18} \\
    Lithuanian & 79.75 & 78.04 & 48.27 & 74.62 & 59.13 & 80.76 & 80.87 & \textbf{86.06} & 71.31 & 75.26 & \textcolor{blue}{83.47} \\
    Polish & 87.65 & 89.17 & 71.52 & 78.20 & 0.0 & 89.78 & \textcolor{blue}{90.66} & \textbf{91.46} & 88.31 & 81.59 & 90.46 \\
    Russian & 90.73 & 92.83 & 78.56 & 90.56 & 66.33 & 92.64 & 93.59 & \textcolor{blue}{94.01} & 90.90 & 79.63 & \textbf{94.09} \\
    Slovak & 87.04 & 89.59 & 64.28 & 86.92 & 67.45 & 89.66 & 90.25 & \textbf{94.96} & 86.05 & 76.42 & \textcolor{blue}{92.73} \\
    Swedish & 83.20 & 85.20 & 67.26 & 81.54 & 63.12 & 88.03 & 86.62 & \textbf{89.90} & 84.91 & 80.98 & \textcolor{blue}{88.64} \\
    Tamil & 52.27 & 39.32 & 42.53 & 58.69 & 0.0 & 59.33 & 58.94 & \textbf{65.58} & 51.73 & 75.44 & \textcolor{blue}{62.06} \\
    Ukrainian & 86.92 & 86.09 & 63.42 & 83.90 & 0.0 & 88.86 & 88.94 & \textbf{92.78} & 87.51 & 77.24 & \textcolor{blue}{90.91} \\
    \hline
    \textbf{Average} & 83.79 & 83.57 & 65.81 & 81.58 & 30.03 & 86.97 & 87.07 & \textbf{89.24} & 83.64 & 79.87 & \textcolor{blue}{88.04} \\
    \hline
    \end{tabular}
    \end{adjustbox}
    \caption{Evaluation scores on the official test data on the language-specific test files submitted by each team.
    We also include the official reference system (\textbf{off. reference}) which copies the gold tree to the enhanced graph as well as (\textbf{our best run}) which is our best post-deadline run, which corresponds to the \textbf{+Concat+MTL} run in Table~\ref{tab:eval}. The first and second top scoring models in each language are specified with black and blue color, respectively.}
    \label{tab:team_eval}
\end{table*}
\subsection{Treebank Concatenation}

In our official system, we used just one treebank per language.
Our next experiment is to
investigate the effect of concatenating all treebanks with enhanced UD annotations
for a language.
We hypothesize that there could be a positive transfer from learning similar (within-language) treebanks and that it would make our parser
more robust to the multiple domains in the test data.
This means that for
Czech we concatenate the PDT, CAC and FicTree treebanks,
for Dutch, Alpino and LassySmall,
for English EWT and GUM, and
for Estonian EDT and EWT.
For Tamil, we concatenate English EWT and GUM training data to Tamil to address
the very poor evaluation score of our official submission, taking inspiration
from \newcite{wang-etal-2020-enhanced} who observe substantial positive effects
when they add Czech and English data to the Tamil treebank.\footnote{We
    did not include Czech to reduce training time.
}
The results are listed in
column [4]
of Table~\ref{tab:eval}.
Treebank concatenation helps for all languages but most notable is the improvement of
over 12 points ELAS
or a relative error reduction of 24\%
for Tamil, the language with the least amount of training data in the task.



\subsection{Joint Learning of Basic and Enhanced Dependency Parsing}

The official
reference
system
submitted by the shared task organizers which copies the gold trees to the enhanced representation
performs very well with 79.87 ELAS (see Table~\ref{tab:team_eval}).
Thus, there is evidence that the
basic tree and enhanced graph contain a lot of mutual information.
Previous methods which have leveraged the basic representation for producing EUD graphs (see Sec.~\ref{sec:related})
have focused on using heuristic rules to convert the basic tree to EUD \citep{schuster-manning-2016-enhanced, ek-bernardy-2020-much, dehouck-etal-2020-efficient},
using the basic tree as input features to the enhanced parsing model \citep{barry-etal-2020-adapt}
or converting the enhanced graph to a richer basic representation \citep{kanerva-etal-2020-turku}.


In our final experiment, we try to leverage the information from the basic tree by
jointly learning to predict the enhanced graph and the basic tree, testing whether
performing basic dependency parsing and EUD parsing in a multitask setup is beneficial
for EUD parsing.
Given the positive effects seen through concatenation,
for those languages where we performed concatenation,
we also train multitask models on the concatenated versions of treebanks.
%
%
We use our EUD parsing model as in Section~\ref{sec:overview} 
and integrate with additional basic dependency parsing component~(as shown in the right part of Figure~\ref{fig:architecture}) which is the biaffine parsing model of \newcite{DBLP:journals/corr/DozatM16} and train both parsers jointly.
The losses of the two components are combined with equal weight.
The results are listed in 
column [5]
of
Table~\ref{tab:eval}.

\paragraph{Single Treebanks}
First, we compare the multitask model to the \texttt{XLM-R\textsubscript{Large}} run for languages where we did not perform concatenation.
Predicting the basic tree and the enhanced graph in a multitask setting yields improvements for all languages, particularly for Arabic, French and Ukrainian.
%

\paragraph{Multitask Model and Treebank Concatenation}
When used alongside treebank concatenation,
multitask learning can help for Dutch, Estonian and Tamil where it provides additional performance gains.
It is interesting to note that concatenation alone is more helpful for Czech and English where we see slight performance drops
and multitask learning is not helpful when trained on concatenated Polish treebanks.

The positive 
contribution
of multitask learning for all languages when not performing treebank concatenation,
could mean that it would be useful in settings
where only one treebank with the enhanced representation is available for a language
and the basic tree could be used as auxiliary information to predict the enhanced representation.

\paragraph{Comparison to Official Systems} Our best unofficial run \texttt{+Concat+MTL} is added to Table~\ref{tab:team_eval}.
Compared to the other official runs,
the ELAS scores of this run ranks in second place for 13/17 languages
and places first for Italian and Russian.

%% file: conclusion.tex
We have described the \textit{DCU-EPFL} submission to the IWPT 2021 Shared Task on Parsing into Enhanced Universal Dependencies.
Our approach uses a single multilingual Transformer encoder as well as an enhanced dependency parsing component.
Our official system placed 6th out of 9 teams.
In post-deadline experiments, we show how our submission can be improved by leveraging better upstream pre-processing,
a larger encoder,
concatenating treebanks
as well as introducing a multitask parser that can parse the basic tree and enhanced graphs jointly.